\pdfoutput=1
\documentclass[11pt]{article}
\usepackage[]{EMNLP2023}
\usepackage{appendix}

\usepackage{times}
\usepackage{latexsym}
\usepackage{graphicx}
\usepackage{subcaption}
\usepackage{array}
\usepackage{amsmath}
\usepackage{booktabs}
\usepackage[T1]{fontenc}

\usepackage[utf8]{inputenc}

\usepackage{microtype}

\usepackage{inconsolata}

\title{\textit{Detecting the Clinical Features of Difficult-to-Treat Depression using Synthetic Data from Large Language Models.}}

\author{Isabelle Lorge\thanks{\> Department of Psychiatry, University of Oxford, UK}, Dan W. Joyce\thanks{\hfill Civic Health Innovation Lab and Institute of Population Health, University of Liverpool, UK} \thanks{\> Mersey Care NHS Foundation Trust, UK} , Niall Taylor\footnotemark[1], Alejo Nevado-Holgado\footnotemark[1], \\ {\bf Andrea Cipriani}\footnotemark[1] \thanks{\> Oxford Health NHS Foundation Trust, Warneford Hospital, UK} \thanks{\> Oxford Precision Psychiatry Lab, NIHR Oxford Health Biomedical Research Centre, UK}, {\bf Andrey Kormilitzin}\footnotemark[1]}

\begin{document}
\thispagestyle{plain}
\pagenumbering{arabic}
\maketitle

\begin{abstract}
Difficult-to-treat depression (DTD) has been proposed as a broader and more  clinically comprehensive perspective on a person's depressive disorder where despite treatment, they continue to experience significant burden. We sought to develop a Large Language Model (LLM)-based tool capable of interrogating routinely-collected, narrative (free-text) electronic health record (EHR) data to locate published prognostic factors that capture the clinical syndrome of DTD.  In this work, we use LLM-generated synthetic data (GPT3.5) and a Non-Maximum Suppression (NMS) algorithm to train a BERT-based span extraction model. The resulting model is then able to extract and label spans related to a variety of relevant positive and negative factors in real clinical data (i.e. spans of text that increase or decrease the likelihood of a patient matching the DTD syndrome).  We show it is possible to obtain good overall performance (0.70 F1 across polarity) on real clinical data on a set of as many as 20 different factors, and high performance (0.85 F1 with 0.95 precision) on a subset of important DTD factors such as history of abuse, family history of affective disorder, illness severity and suicidality by training the model exclusively on synthetic data.  Our results show promise for future healthcare applications especially in applications where traditionally, highly confidential medical data and human-expert annotation would normally be required. 

\end{abstract}

\section{Introduction}\label{introduction}
Major depressive disorder (MDD) is highly prevalent with a heavy economic,  social and personal burden of disability worldwide and affecting up to 6-12\% of the adult global population, a proportion which has been rising in the last few years \cite{santomauro2021global}. The DSM-V (Diagnostic and Statistical Manual of Mental Disorders V) operationalises MDD as a continuous period of at least two weeks characterised by a change in mood leading to loss of interest or pleasure in activities, along with other symptoms such as weight loss or gain, sleep or cognitive issues and suicidal thoughts causing substantial distress or impairment to functioning \cite{american2013diagnostic}. 

Depressive disorders are conditions with sub-optimal treatment outcomes, with up to 70\% of patients failing to achieve remission after receiving pharmacological treatment \cite{caldiroli2021augmentative}. This led to the development of the concept of treatment-resistant depression (TRD) and although definitions vary \cite{brown2019current} a common point of consensus is the failure to achieve treatment response after sequential, adequate-duration and minimally-effective dosed trials of two antidepressant-class medications.  Designating a patient as having TRD focuses on acute-phase symptom improvement in response to pharmacological intervention.  

Despite these efforts at refining the definition of depression to address treatment responsiveness, the cumulative rate of chronicity and lack of response or remission amongst MDD patients remains high, with around 30\% of patients not achieving remission even after four courses of antidepressants \cite{rush2006acute}.  In addition, treatment resistant depression may be associated with higher risks of suicide \cite{papakostas2003hopelessness}. This underlines the importance of improving the identification of the relevant features (or signature) of people with depression where treatment has not provided adequate remission. 

The relatively new concept of difficult-to-treat depression is proposed \citet{mcallister2020identification} as a more comprehensive model that emphasises biomedical, psychological and social factors and interventions that may influence response to treatment beyond the acute symptomatic response to pharmacological interventions in TRD.  

\section{Machine learning models for treatment outcome prediction}
A number of studies have examined the use of machine learning (ML) techniques to directly identify treatment response \cite{perlis2013clinical,nie2018predictive}. Several of these re-use the Sequenced Treatment Alternatives to Relieve Depression (STAR*D) cohort dataset \cite{rush2006acute}, where four successive treatment steps were administered, patients having the option of exiting the study if they experienced sufficient remission of symptoms after a given step.  These ML studies identified predictive features including depression severity, the presence of co-morbid physical illness, psychosis, Post-Traumatic Stress Disorder (PTSD), anxiety disorder alongside minority ethnic heritage, work and poor social adjustment. Other factors identified in later studies include recurrence of depressive episodes, age, response to a first antidepressant, suicidality, educational attainment and occupational status \cite{kautzky2017refining, kautzky2019clinical}.  Another recent study uses age, gender, race, diagnostic codes (including both ICD-9 and ICD-10), current procedural terminology codes and medications with a tree-based algorithm \cite{lage2022efficiently}. These studies achieved relatively good performance using traditional ML models such as random forests, GBDT (gradient-boosted decision trees) and logistic regression, however they use data from structured data fields in electronic health records (EHRs) and focus on delivering a binary TRD outcome.

In addition to the above studies using structured data fields, a few studies have used terms extracted from narrative EHR notes, notably \citet{perlis2012using}, who leverage regular expressions to extract terms identified by clinicians as important for prediction and compare the outcome of a model trained with billing data to a logistic regression model trained with concepts extracted from narrative notes, finding that the latter performs substantially better. Similarly, \citet{sheu2023ai} use a variety of deep learning models with features from both categorical variables and terms extracted from narrative notes using regular expressions (pattern matching) as well as Longformer vectors of the clinical note history to predict treatment outcome.  

Thus, previous works leveraged either structured categorical data (e.g., registered comorbidity, socio-demographic factors etc.), overall scores on standardised questionnaires, terms extracted from notes using rule-based techniques such as regular expressions or, in the case of \citet{sheu2023ai}, a non-explainable vectorised representation of the full patient history.

An alternative is to use Natural Language Processing (NLP) models trained to extract information from the narrative clinical notes \cite{vaci2020statistics, kormilitzin2021med7}. For example, recent work by \citet{kulkarni2024enhancing} uses a BERT model for extracting suicidality and anhedonia in mental health records using ground-truth manually annotated EHR data. 

NLP models are both flexible and ecologically valid, and have been increasingly used successfully for various applications in psychiatry \cite{vaci2021real, senior2020identifying} and clinical neuroscience in recent years \cite{le2021machine, liu2022personalised}. The models can be trained either to directly predict a phenotype based on treatment outcome using text features as in \citet{sheu2023ai} or, as in the current study, to extract relevant features suggestive of a phenotype or syndrome (e.g., in the case of DTD, a history of abuse, self-harm and suicidality) to present to clinicians for decision making.  However, there are significant hurdles to obtaining training data, such as data scarcity and the high financial and time cost of manual annotations. 

\section{Difficult-to-treat depression}
Difficult-to-treat depression or DTD is a more recent framework than TRD which was developed following a number of issues becoming apparent with using the latter concept. As described in \citet{mcallister2020identification}:
\begin{itemize}
    \item The current definition of TRD largely ignores psychotherapy and neurostimulation treatments
    \item The definition does not allow for differential levels of success in response or remission
    \item The phrasing could imply treatment failure is a property of the patient, rather than inadequacy of the intervention
    \item The term implies a medical model which may exclude social and environmental factors previously shown to be significant predictors of treatment response
\end{itemize}

For these reasons, the more inclusive and flexible concept of difficult-to-treat depression was put forward by an international consensus group and a number of factors were identified through literature review and expert consensus, grouped under the categories of PATIENT, ILLNESS and TREATMENT factors \cite{mcallister2020identification}. Given the novelty of the framework, there have been few attempts to operationalise it, with the exception of \citet{costa2022burden} who partitioned patients from five specialist mental health National Health Service (NHS) Trusts according to a criterion encompassing both recurrence and resistance (at least 4 unsuccessful treatments including two antidepressant medications) and analysed correlations with environmental and clinical factors, confirming previous findings. In the current work, we use the concept of difficult-to-treat depression rather than treatment-resistant depression.

\section{Large Language Models and synthetic data for medical and mental health research}
The recent development of large language models (LLMs) with substantially increased size and capabilities allowed great strides of improvement across domains such as question answering and text summarisation, including in the biomedical domain \cite{agrawal2022large, hu2023zero, liu2023deid, tang2023evaluating,taylor2023clinical}; becoming potentially able to perform information extraction on large quantities of data. However, privacy considerations prevent directly feeding highly-confidential patient data into LLM APIs such as OpenAI's chatGPT \cite{brown2020language}. In addition, while LLMs show impressive performance in applications related to text generation, they still fall short of specifically trained SOTA systems on biomedical NLP tasks \cite{ateia2023chatgpt}. A recent paper demonstrated that even when fine-tuned on target tasks, a LLaMA model, \cite{touvron2023llama} orders of magnitude bigger (7B vs 100M) and requiring substantially more compute power still underperforms compared to BERT-based discriminative models by up to 10\% \cite{yang2023mentallama}. For these reasons, a paradigm is emerging whereby a smaller local domain or task-specific model is fine-tuned on synthetic labelled data generated by an LLM, mitigating concerns of privacy as well as efficiency. 

This paradigm has been used successfully for diagnosis and mortality prediction by  \citet{kweon2023publicly}, who generated synthetic clinical notes using GPT3.5-turbo \cite{brown2020language} and trained a domain specific LlaMA model, and for Named Entity Recognition and Relation Extraction by \citet{tang2023does}, who used prompts involving named entities and gene/disease relations from PubMed and fine-tuned BERT family models. Another recent work uses LLM-generated synthetic data to augment gold annotated data training a Flan-T5 model to perform multilabelling of sentences for social determinants of health (SDoH) such as housing, employment, transportation, parental status, relationship status and social support \cite{guevara2024}. However, perhaps due to the small amount of added synthetic data (1800 synthetic sentences added to over 30k gold sentences), the synthetic addition does not lead to substantial or consistent improvement in performance  (in fact it sometimes worsens it) and the performance training with synthetic data only is extremely poor (< 0.1 F1). In the domain of psychiatry, another study augmented training data with LLM-generated interviews and used traditional machine learning classifiers for binary classification of Post-Traumatic Disorder (PTSD) with a 10\% increase in performance \cite{wu2023automatic}.

The generated synthetic data has the potential to mimic important statistical properties and patterns of real data while avoiding the expensive and effortful process of obtaining large quantities of labelled data. Therefore, we intend to attempt training a DTD feature span extraction model exclusively on synthetic data obtained from LLMs. 


\section{Aims}
Previous studies used machine learning techniques to predict treatment response or identify treatment-resistant depression. In contrast, the more comprehensive concept of difficult-to-treat depression allows us to leverage a wider variety of factors -- rather than relying on a strict `two-course' acute-phase response to pharmacological interventions -- that potentially enables early detection and more personalised care that addresses reasons for sub-optimal treatment response. Recent research has indeed been focusing on early detection and linking prognostic factors with a continuum of treatment response \cite{lage2022efficiently, sheu2023ai}.


Furthermore, previous works mostly focused on sentence classification rather than the more complex process of span extraction. Contrary to multilabelling, span extraction presents clinicians with the specific part of the text which is linked to a particular label, allowing them to focus more efficiently on relevant information. Finally, to our knowledge no previous study successfully trained a model exclusively on synthetic data for the purpose of extracting prognostic factors. Success, even partial, would represent an extremely important step in developing AI applications for healthcare, given the known issues of data scarcity and manual labelling costs. 


Therefore, the present paper aims to:
\begin{itemize}
    \item Add to the recent body of works demonstrating the utility of leveraging synthetic data for decision-supporting information extraction in biomedical domains
    \item Introduce an annotation scheme (abductive annotation) that leverages domain experts on narrative clinical notes to facilitate explicit extraction of PATIENT, ILLNESS and TREATMENT related-factors for DTD 
    \item Build a curated synthetic dataset of annotated narrative clinical notes which can be freely shared with the research community
    \item Build and train a model which extracts spans of text and labels them with the relevant factor with the goal of downstream clinician decision guidance \footnote{The code and synthetic data are available at \url{https://github.com/isabellelorge/dtd}}
\end{itemize}

\section{The DTD Phenotype}
In consultation with a clinician expert, we operationalise the prognostic factors originally reported in \citet{mcallister2020identification} into the annotation schema presented in Table \ref{counts}. There are several reasons for the changes made. First, we aim to develop a schema which could potentially be extended to other conditions or strata of populations (e.g. we created general categories for physical and mental comorbidities). Second, because we leverage the use of a large language model for generating and annotating data, we endeavour to create labels which are semantically transparent enough that they are likely to be understood by the model and yield better annotation accuracy.  

Finally, When a clinician interprets the data contained in a patient's EHR (e.g., to establish a diagnosis) they will employ abductive (rather than inductive or deductive) reasoning \cite{sep-abduction, rapezzi2005white, altable2012logic, reggia1985answer} meaning they will seek evidence to support or refute a number of competing hypotheses (e.g., differential diagnoses).  Consequently, we wish to be able to extract both \textit{positive} and \textit{negative} evidence in favour or against a designation of difficult-to-treat depression in order to provide clinicians with a comprehensive picture of the patient's current presentation, illness and treatment history for downstream decision making.  


\begin{table}[h!]
\small
\begin{tabular}{|m{4.8cm}|m{0.65cm}| m{0.65cm}|} 
 \hline
 \textbf{Factor} &  \textbf{Orig}. & \textbf{New}\\ 
 \hline
 \smallskip
    NO\_ANNOTATION & 10708 & 35884 \\
 family\_member\_mental\_disorder\_POS & 1188 & 2337 \\
 childhood\_abuse\_POS & 1052 & 2116\\
 non\_adherence\_POS & 954 & 1919 \\
 side\_effects\_POS & 883 & 1768\\
 recurrent\_episodes\_POS & 876 & 1774 \\
 multiple\_antidepressants\_POS & 862 & 1744 \\
 multiple\_psychotherapies\_POS & 860 & 1726\\
 physical\_comorbidity\_POS & 824 & 1673\\
 long\_illness\_duration\_POS & 796 & 1603 \\
 severe\_illness\_POS & 732 & 1447\\
 anhedonia\_POS & 731 & 1449\\
 suicidality\_POS & 706 & 1396\\
 antidepressant\_dosage\_increase\_POS & 705 & 1420 \\
 multiple\_hospitalizations\_POS & 668 & 1345\\
 older\_age\_POS & 545 & 1138\\
 mental\_comorbidity\_POS & 473 & 941 \\
 improvement\_POS & 403 & 786\\
 substance\_abuse\_POS & 400 & 787\\
 illness\_early\_onset\_POS & 382 & 751\\
 substance\_abuse\_NEG & 314 & 787\\
 multiple\_hospitalizations\_NEG & 247 & 1375 \\
 suicidality\_NEG & 160 & 1178 \\
 older\_age\_NEG & 140 & 960 \\
 physical\_comorbidity\_NEG & 135 & 1133 \\
 abuse\_POS & 94 & 187\\
 improvement\_NEG & 78 & 721\\
 mental\_comorbidity\_NEG & 64 & 680\\
 non\_adherence\_NEG & 60 & 709 \\
 abuse\_NEG & 53 & 690\\
 childhood\_abuse\_NEG & 42 & 805\\
 family\_member\_mental\_disorder\_NEG & 34 & 996\\
 side\_effects\_NEG & 29 & 701\\
 antidepressant\_dosage\_increase\_NEG & 18 & 506 \\
 multiple\_psychotherapies\_NEG & 16 & 746\\ 
 multiple\_antidepressants\_NEG & 11 & 587\\
 long\_illness\_duration\_NEG & 7 & 567\\
 severe\_illness\_NEG & 6 & 488 \\
 recurrent\_episodes\_NEG & 6 & 718 \\
 illness\_early\_onset\_NEG &  5 & 382\\
 anhedonia\_NEG & 4 & 541\\
 \hline
\end{tabular}
\caption{Annotation schema labels and span counts for original and new (final) datasets. We shorten the polarity words in the table for space considerations (the labels used in prompts and GPT3.5-turbo outputs use full words, e.g., [PATIENT\_FACTOR(POSITIVE):older\_age]).}
\label{counts}
\end{table}

\section{Synthetic dataset}
In the first instance, we prompt ChatGPT (GPT3.5-turbo-0613) through its Python API to generate and annotate a dataset of 1000 clinical notes with labels from our annotation schema (e.g., [PATIENT\_FACTOR(POSITIVE):older\_age]). We used a temperature of 1.2 and default values for remaining parameters. 
Experimenting with various prompts, we find that the best prompts balance the need to provide examples to correct recurrent mistakes in model's behaviour against the tendency for the model to frequently repeat given examples, even with a high temperature parameter to encourage diversity in token generation. We thus add examples as needed to correct various errors (tendency to not label age, to output only positive labels, to group all labels after the sentence rather than after the relevant span, etc.). An example generated annotated note can be found in Appendix \ref{AppendixA} and the final prompt used can be found in Appendix \ref{AppendixB}. 
We notice from manual inspection of examples that the output is relatively accurate, though the model does output errors, meaning that the labels are noisy and should be considered as a silver rather than gold standard.

We use a combination of regular expressions and heuristics to extract labels from each sentence, discarding labels which do not fit our schema (in a small proportion of notes the model hallucinates new factors or uses a different formatting). We first experiment with a syntactic heuristic to extract shorter spans (labelling the closest finite verb phrase with each label) but eventually settle on a simpler heuristic of labelling all text before each label up to sentence boundary (e.g., for \textit{'XXXXX [label\_X] YYYYY [label\_Y]'}, we extract \textit{XXXXX} as span for label X and \textit{YYYYY} for label Y). This is because early prompt attempts showed that when prompting chatGPT for span boundaries (e.g., \textit{'YYYYY\{XXX\} [label\_X]'}), the boundaries obtained were very error-prone and unreliable, while a strategy of simply asking the model to insert the label after each relevant span yielded much better results. 

From the span counts in Table \ref{counts}, it can be seen that the label distribution is heavily skewed, with a very strong bias in favour of positive factors against negative factors. We believe this would reflect clinicians' annotations, as positive evidence is much more likely to be expressed and noticed than negative evidence and arguably contributes more heavily to final decision making. However, to reduce data imbalance we prompt the model for another 1000 notes with the same prompt and a third set of 1000 notes with a prompt asking exclusively for negative labels. The final dataset thus contains 3000 notes which yield 75094 sentences. The updated factor counts of the final dataset can be seen in Table \ref{counts} and the negative prompt can be found in Appendix \ref{AppendixC}. 

We explore the word distribution for each factor to try and get a sense of the synthetic dataset's diversity and/or repetitiveness. To achieve this, we extract words from labelled spans for each label and calculate their TF-IDF score, defined as:
\begin{equation}
tfidf\_score = (w_{l}/W_{l})/\log(w/w_{l} + 1)
\end{equation}

Where $w_{l}$ is the frequency of a given word for a given label, $W_{l}$ is the total number of words for the label and $w$ is the total frequency for the word across labels. A sample of labels and their 10 highest scoring words can be seen in Table \ref{tfidf}, along with their prevalence (number of occurrences/number of label spans). 
Interestingly, while there is a tendency for the model to repeat examples given in the prompt, the extent of this behaviour varies substantially depending on specific examples. Indeed, while our prompt example for \textit{family\_member\_mental\_disorder\_POSITIVE} mentions anxiety, the most frequent disorder which appears for this label is bipolar disorder (48\% of spans), whereas an overwhelming 80\% of spans labelled with \textit{mental\_comorbidity\_POSITIVE} do include the word `anxiety' without having been prompted with it. Similarly, our prompt example for \textit{childhood\_abuse\_POSITIVE} mentions motherly abuse but only 19\% of spans with this label contain the word `mother', superseded by 34\% of spans mentioning the word `father'. 
The \textit{physical\_comorbidity\_POSITIVE} factor is more evenly distributed than \textit{mental\_
comorbidity\_POSITIVE}, with most frequent conditions split between hypertension, diabetes and fibromyalgia. The model displays strong biases which are not driven by prompt examples, as evidenced by 70\% of \textit{substance\_abuse\_POSITIVE} spans mentioning the word `alcohol'. We hypothesise that the above biases result from the distribution of the model's training data. 

We also note that spans from some labels have a high probability of mentioning all label words explicitly: \textit{abuse}, both NEGATIVE (91\%) and POSITIVE (80\%), \textit{side\_effects}, both POSITIVE (88\%) and NEGATIVE (84\%), \textit{improvement}, both POSITIVE (84\%) and NEGATIVE (73\%), \textit{substance\_abuse\_NEGATIVE} (84\%) and \textit{childhood\_abuse}, both NEGATIVE (76\%) and POSITIVE (75\%). There is a tendency for the model to mention label words explicitly more often for negative than positive labels, probably due to higher variety for spans providing positive evidence. Only 10\% of spans for \textit{physical\_comorbidity\_POSITIVE} and \textit{mental\_comorbidity\_POSITIVE} mention label words explicitly, suggesting that the model still relies on its training data for phrasing and does not systematically repeat prompt labels. The numbers for all labels can be seen in Figure \ref{explicit} in Appendix \ref{AppendixD}.

Finally, to assess the similarity in word overlap between pairs of spans for a given label we calculate the Jaccard similarity between lemmatized words of each pair of spans within a label. The similarities range from 0.02 for spans with no label to 0.35 for \textit{substance\_abuse\_NEGATIVE}. The full numbers for all labels can be seen in Figure \ref{jaccard} in Appendix \ref{AppendixD}.

\begin{table}[h!]
\centering
\begin{tabular}{|m{1.4cm}|m{1.7cm}|m{0.6cm}|}
 \hline
 \textbf{Factor} &  \textbf{Words} & \textbf{\%} \\ 
 \hline
family member mental disorder & 
\small{family \newline bipolar \newline history \newline
disorder \newline mental \newline mother \newline illness \newline diagnosed \newline reports \newline sister} & \small{
0.72 \newline 0.48 \newline 0.87 \newline 0.65 \newline 0.68 \newline 0.44 \newline 0.65 \newline 0.39 \newline 0.41 \newline 0.16} \\
\hline
 childhood abuse & \small{
childhood \newline abuse \newline emotional\newline neglect\newline father\newline history\newline
physical\newline experienced\newline discloses\newline
mother} & \small {
0.83 \newline 0.95 \newline 0.56 \newline 0.30 \newline 0.34 \newline  0.46\newline 0.35\newline 0.22\newline 0.11\newline 0.19}\\ 
\hline
physical comorbidity & \small{
hypertension\newline physical\newline diabetes\newline
comorbidities\newline pain\newline chronic\newline fibromyalgia\newline medical\newline comorbidity\newline
history}  & \small{0.38\newline 0.62\newline 0.29\newline 0.31\newline 0.19\newline 0.22\newline 0.14\newline 0.25\newline 0.24\newline 0.41
}\\
\hline
mental comorbidity & \small{
anxiety\newline generalized\newline disorder\newline comorbid\newline also\newline mental\newline panic\newline diagnosis\newline diagnosed\newline comorbidity}
& \small{
0.80\newline 0.47\newline 0.77\newline 0.28\newline 0.31\newline 0.23\newline 0.08\newline 0.14\newline 0.16\newline 0.15}\\
\hline
substance abuse & \small{
alcohol\newline substance\newline abuse\newline history\newline use\newline mechanism\newline patient\newline specifically\newline cope\newline
also} 
& \small{
0.70\newline 0.82\newline 0.85\newline 0.53\newline 0.20\newline 0.10\newline 0.34\newline 0.17\newline 0.10\newline 0.23}\\
\hline

\end{tabular}
\caption{Sample factors with their 10 highest TFIDF scoring words and their \% or prevalence (n occurrences/n spans).}
\label{tfidf}
\end{table}

\section{Real-World clinical data}

To evaluate the performance of the developed model on real clinical data, we utilised a sample of de-identified secondary mental health records from the Oxford Health NHS Foundation Trust, which provides mental healthcare services to approximately 1.2 million individuals across all ages in Oxfordshire and Buckinghamshire in England. Access to the de-identified data was obtained through the Clinical Record Interactive Search (CRIS) system powered by Akrivia Health, which enables searching and extraction of de-identified clinical case notes across 17 National Health Service Mental Health Trusts in England. For this study, we sampled clinical summaries for 100 adult patients over 18 years old, randomly selected from 19,921 patients with confirmed diagnosis of depression (ICD-10 codes F32 and F33) readily available from structured data fields in CRIS. 

Access to and use of de-identified patient records from the CRIS platform has been granted exemption by the NHS Health Research Authority for research reuse of routinely collected clinical data. The project was reviewed and approved by the Oversight Committee of the Oxford Health NHS Foundation Trust and the Research and Development Team.

\section{Task}
We extract character indexes of start, end and corresponding label for labelled spans in each sentence and remove label text from sentence text. There are 41 labels (positive and negative polarity label for each factor and a 'NO\_ANNOTATION' label).  
\section{Models}
\begin{figure}[h!]
\includegraphics[scale=0.45]{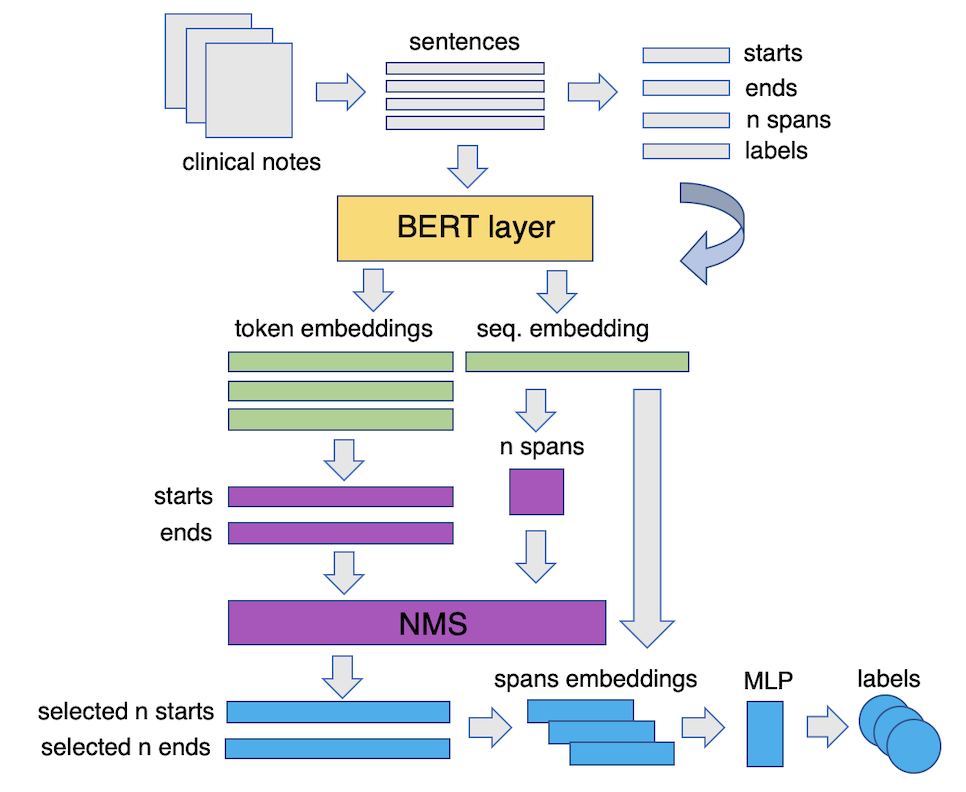}
\caption{Span-level model architecture.}
\label{arch}
\end{figure}
We experiment with a variety of models trading off complexity and granularity. 
\subsection{Token-level}
The first model simply leverages a BERT (base, cased) layer for classification at the token level, thus spans are represented as contiguous series of tokens labelled with a specific label. We use one linear layer and no IOB flags (given that each span starts where the previous span ends). The output of the classification layer for each token is fed to a softmax activation layer which outputs probabilities for each label. This is similar to the traditional technique used for Named Entity Recognition (NER). We use a custom Pytorch class and Huggingface's transformers library  \cite{DBLP:journals/corr/abs-1910-03771} implementation for the BERT layer.
\subsection{Span-level}
Given the longer lengths of our spans compared to traditional NER tasks, we also experiment with a model inspired from Question Answering (QA) systems, where the model does not predict a label for each token in the sequence but instead predicts a `start' and `end' position. However, traditional Question Answering systems are generally limited to one response span, restricting the scope of the classification task, while in our case there can potentially be any number of spans within a given sequence, each of which could be of any length, expanding the training search space exponentially. To solve this issue, we follow \citet{DBLP:journals/corr/abs-1908-05514} and take inspiration from computer vision by using a variant of Non-Maximum Suppression (NMS), whereby non-overlapping outputs are selected in decreasing order of confidence. To achieve this, a separate classifier is trained to predict the number (N) of spans within a sequence from the sequence output of a BERT (base, cased) layer, and subsequently the top N non-overlapping start/end pairs with the highest combined probabilities are selected. We find that given start and end of sentences consistently have very high probabilities for our dataset (since spans are full sentences in cases where there is a single factor in the sentence), a greedy approach on combined start/end probabilities as used by \citet{DBLP:journals/corr/abs-1908-05514}
does not work. Instead, we predict the number (N) of spans and separately select the top N starts and N ends with highest probabilities, which we order by token number so that we select non-overlapping start/end combinations in order by taking the closest end for each start (e.g., if 2 predicted spans with top starts [1, 10] and top ends [20, 10], the indexes are ordered so that the first span is not [1, 20] but [1, 10], and the second span is [10, 20]).
The model goes through each selected start/end pair and uses a linear layer span classifier to label the relevant tokens from the sequence. For training (given possible inconsistencies between true and predicted number of spans), the output of the span classifier is passed through a softmax layer and all predicted spans are max pooled into a single vector to match the ground truth multilabel one-hot encoded vector for the full sentence (soft selection). At inference time, the indices of the maximum probability for each predicted span are taken as labels (hard selection). We also experiment with a version of the model where factors are merged together across polarities and a separate classifier is trained to predict negative polarity from a concatenation of the BERT sequence output and predicted labels \footnote{The results are substantially worse and in the interest of space considerations we do not report them.}.  As for the token-level model, we implement a custom class and loss function in Pytorch and leverage Huggingface's implementation of the BERT layer. The architecture of the span-level model can be seen in Figure \ref{arch}.

\subsection{Sentence-level}
Finally, we also model the task in a simplified way as a multilabel sentence classification task. For this, we again leverage the sequence output of a BERT (base, cased) layer which we feed to a linear classifier layer. The output of the classifier is then passed through a sigmoid activation layer and labels with probabilities above a 0.5 threshold are taken as predictions. In this case the start and end positions of spans for the predicted labels are unknown. Again, we use Huggingface's BERT layer implementation and a custom Pytorch class for classification with one linear layer.

\section{Training}
We split the synthetic dataset into train, development and test sets with proportions 0.8, 0.1 and 0.1. We use a batch size of 16, learning rate of 3e-05, dropout rate of 0.1 and weight decay of 0.001. We experiment with different forms of class weights to mitigate the imbalance in labels. For the span model, we find the model performs better with a logarithmically scaled class weighted loss, whereas there is no difference for the token level model, and the sentence-level model performs better without class weights. We train until convergence, 4 epochs for span, 5 epochs for the token model and 7 epochs for the sentence-level model.

\begin{figure*}
\hspace{1cm}
\centering
\includegraphics[scale=1]{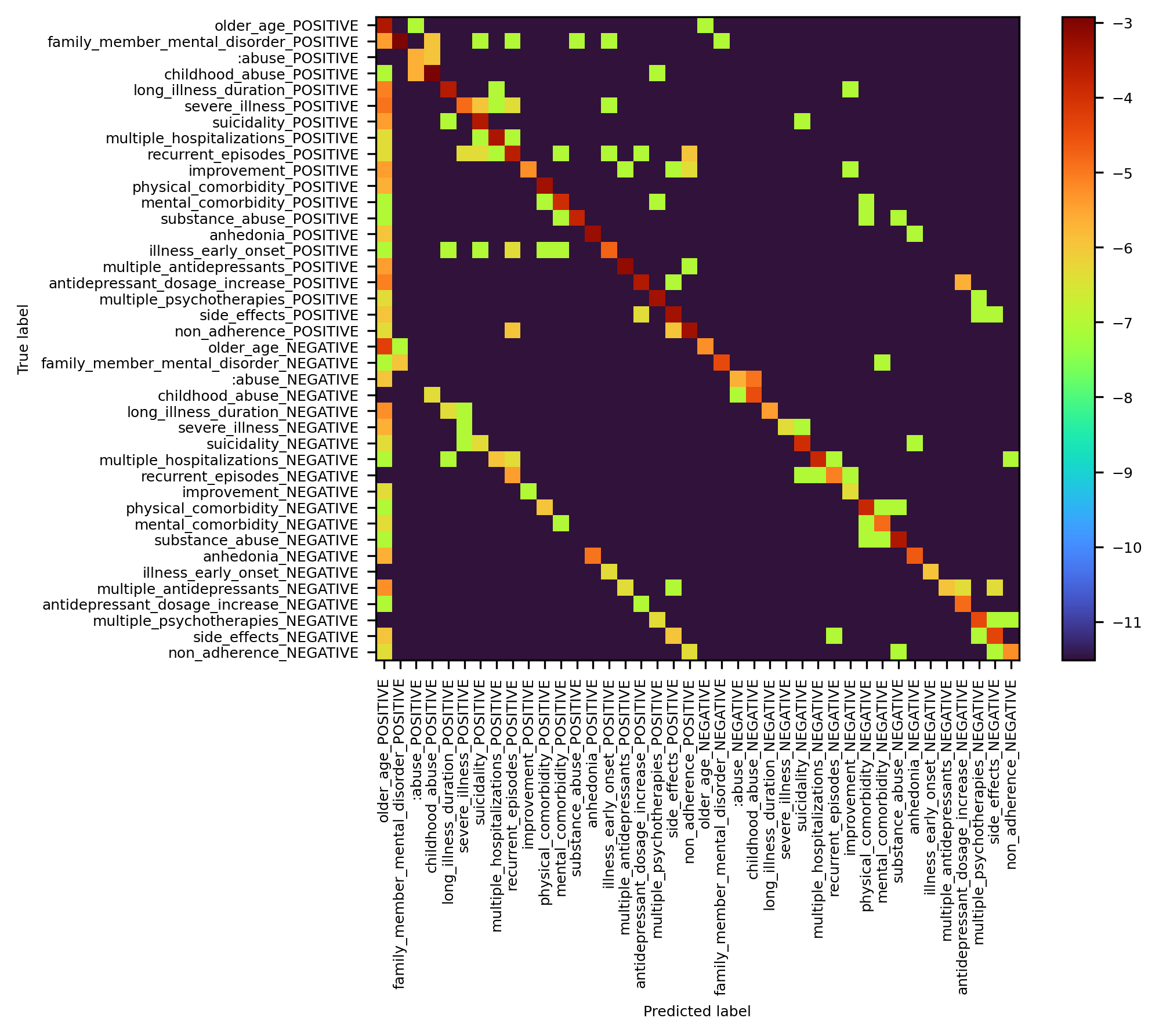}
\caption{Log-scaled confusion matrix (synthetic data).}
\label{conf_synth}
\end{figure*}


\section{Experiments}
\subsection{Synthetic data}
We first test the model on a held-out test set of synthetic data. 
\subsubsection{Results}
The results on the synthetic test set for the three models can be seen in table \ref{results} and a log-scaled confusion matrix can be seen in Figure \ref{conf_synth}
\footnote{While we used simplified terms in our labels to ensure the best LLM output after prompting trial and error, the appropriate medical terms should be used in the final output of the model for the following: \textit{episodes severity} for \textit{severe illness}, \textit{episode remission} for \textit{improvement}, \textit{medical comorbidity} for \textit{physical comorbidity}, \textit{substance use} for \textit{substance abuse}, \textit{adequate dosage} for \textit{antidepressant dosage increase} and \textit{adherence medication} for \textit{non-adherence}.}. 
Given the imbalance in labels, we present precision, recall and F1 for each class as well as F1 averaged across classes, with macro averaged F1 being the accepted standard metric in similar tasks. The three models perform fairly well. We find our custom span-level model outperforms the simpler, token-level model by a large margin (0.65 vs 0.57). To our surprise, we also find that our span level model slightly outperforms the sentence-level classification model, despite the added difficulty of having to predict the correct start and end of spans in order to be able to accurately predict labels.  The models tend to perform substantially better on the positive classes. 

The sentence and span level models perform similarly across factors, each claiming best performance for an approximately equal number of factors, whereas the token-level model only outperforms other models for the abuse\_POSITIVE class. There appears to be little difference between the performance for the different factor domains (PATIENT, ILLNESS and TREATMENT).

We notice that the model struggles with the category `older age', which points to a well-known shortcoming of language models (along with negation) regarding ability to count and evaluate quantities. The confusion matrix in Figure \ref{conf_synth} shows how the category gets confused with most other classes (lit up vertical line). The worst performing positive classes is \textit{abuse\_POSITIVE} (most likely due to scarcity of training examples) and worst performing negative class is \textit{improvement\_NEGATIVE}. Examination of synthetic data reveals GPT3.5 particularly struggled with correctly labelling negative examples of improvement, which led to many being in fact positive examples (e.g., \textit{There has been improvement in symptoms [ILLNESS\_FACTOR:improvement(NEGATIVE)]}). This seems in fact to be a side-effect of GPT3.5's `common sense', indeed all other negative classes indicate a positive outlook (e.g., no history of abuse or disorder, no substance use, no hospitalisations, etc.), thus when prompted for negative examples the LLM produced sentences coherent with the rest of the note, i.e., sentences mentioning improvement. The secondary lit up diagonal on the left of the confusion matrix confirms that the model struggles with negation, with a tendency to predict the positive counterpart for negative classes and vice versa (right-hand side diagonal).

\subsection{Clinical data}
While performance of the span extraction model is well above chance, it still appears low given that we are testing the model within distribution (i.e., on the same synthetic data it was trained with) with the expectation of a drop in performance when taken out-of-distribution (i.e., to EHR clinical data). Grounds for this concern are confirmed by a first test on a test set of 482 sentences from the above mentioned electronic health records annotated by a consultant psychiatrist for our target features, which yields quite a low overall performance (0.30 F1). This prompts us to further examine the synthetic training data and to perform in-depth error analysis to understand the reasons underlying the low performance both on the synthetic and on the clinical test sets. The analysis reveals the following issues:
\begin{itemize}
    \item Many spans are mislabelled across labels
    \item Many spans tend to be repetitive/reuse the same words for a given label (e.g., `family' for family disorder, `abuse' for substance abuse, etc.)
    \item The style and format of the synthetic notes (articulate, using possessives, articles, pronouns, auxiliary verbs, punctuation and connectors consistently) differs substantially from that of the real clinical data (telegraphic, with frequent ellipsis of articles, pronouns, verbs and connectors, deidentification placeholders, missing spaces and punctuation, inconsistent casing, etc.) which leads to lack of robustness in the model predictions (i.e., predictions changing when a pronoun is changed, period is removed, etc.)
    \item In up to 30\% of cases (from manually examining 100 spans), spans with negative labels are in fact positive spans, which means 1) the model is confused between negative and positive factors and fails to properly learn them and 2) many `errors' in the synthetic test set are in fact correct predictions with incorrect ground truth labels 
\end{itemize}
To address these challenges, we thus perform the following changes:
\begin{itemize}
    \item We use heuristics to remove the bulk of wrongly labelled spans (by removing most spans for a given label which contain keywords from other labels, leaving a few to avoid overfitting)
    \item For each label, we upsample the `diverse' spans (i.e., spans that do not mention label words explicitly, e.g., spans that do include `family' or `disorder' for family disorder, or `sustance` and `abuse' for substance abuse)
    \item We add `noise' to the data by randomly removing possessives, articles, pronouns, auxiliary verbs and punctuation and replacing some pronouns with deidentification placeholder strings (``FFFFF'' and ``XXXXX'') that are used to pseudonymise our research EHR data
    \item We switch to BERT-base-uncased to avoid casing issues
    \item We remove the older age category (which is available as structured data in an EHR) and merge childhood abuse and abuse classes into a single abuse category (due to the scarcity of the latter class)
    \item We remove the sentences with no labels from the original dataset and specifically prompt GPT3.5-turbo for 3000 clinical sentences which do not mention our target features. This is done both because the original unlabelled sentences had a high amount of noise (the plan would mention prognostic/risk factors or features but was not annotated by the language model) and to ensure more diversity in unlabelled sentences.
\end{itemize}

The resulting dataset contains 55924 sentences. We retrain our model on this new synthetic data and obtain an average F1 of 0.75 on the synthetic test set (keeping in mind that since a percentage of test set examples are wrongly labelled, this does not reflect the real performance of the model). We then re-test the model on the annotated EHR sentences. 

\subsubsection{Results}
The results on the clinical data can be seen in Table \ref{results_real}. The overall performance is 0.60 F1, with manual perturbation analysis indicating that the model is much more robust to changes in style or format (e.g., removing pronouns or punctuation). The performance varies widely across classes, with some classes performing very well (0.95 F1 for \textit{recurrent episodes POSITIVE}) while others show poor performance (0.19 F1 for \textit{non-adherence NEGATIVE}). Performance for negative classes is again unsurprisingly worse than for positive classes. In general, recall is higher than precision, with a tendency for the model to overpredict factors. This is unsuprising given the high overlap in topic and broad range of our factors. Indeed, negative classes such as negative multiple antidepressants (e.g., \textit{She takes antidepressant X}), negative anhedonia (e.g., \textit{She enjoys hobbies like painting and reading} or negative abuse \textit{e.g., \textit{She likes spending time with her family}} might be virtually indistinguishable from \textit{NO\_ANNOTATION} spans, and arguably most sentences could be said to contain a span within our target factors if we include negative classes. We come back to this issue in the Discussion.

To remedy the model's oversensitivity, we can increase the confidence threshold for predicting factors, that is only outputting factor predictions above 0.5 probability and no label otherwise. This also has the advantage that it ensures any output predictions are robust (i.e., made with high confidence rather than being `lucky guesses'), which is particularly desirable in medical settings. Finally, it increases recall of the \textit{NO\_ANNOTATION} class to 0.90. When this threshold is increased to 0.5, four high confidence classes emerge (\textit{abuse POSITIVE, family member mental disorder POSITIVE, severe illness POSITIVE} and \textit{suicidality POSITIVE}) which can be confidently predicted with 0.85 average F1 and 0.95 precision (see Table \ref{results_highconf}.)

A significant factor contributing to the model's lower performance is the model confusing positive and negative classes (partially due to noise in the synthetic training data as mentioned previously). This is demonstrated by collapsing predictions across polarities, which increases average F1 to 0.70 (see Table \ref{results_nonpol}). Given the goal of the model is to present clinicians with evidence from extracted spans for their consideration, the non-polarised model could also have clinical utility. 

\begin{table*}[htbp]
\tiny
\centering
\begin{tabular}{lrrrrrrrrr}
\toprule
\multicolumn{1}{c}{}& \multicolumn{3}{c}{\textbf{Sentence-level model}} & \multicolumn{3}{c}{\textbf{Span-level model}}& \multicolumn{3}{c}{\textbf{Token-level model}} \\
\midrule
Class & Precision & Recall & F1 Score & Precision & Recall & F1 Score & Precision & Recall & F1 Score  \\
\midrule
older age POSITIVE & 0.46 & 0.57 & 0.51 & 0.42 & 0.83 & \textbf{0.56} & 0.5 & 0.57 & 0.54 \\
family member mental disorder POSITIVE & 0.75 & 0.9 & \textbf{0.82} & 0.74 & 0.9 & 0.81 & 0.7 & 0.89 & 0.78 \\
abuse POSITIVE & 0.6 & 0.25 & 0.35 & 0.69 & 0.38 & 0.49 & 0.84 & 0.4 & \textbf{0.54} \\
childhood abuse POSITIVE & 0.77 & 0.87 & \textbf{0.81} & 0.73 & 0.89 & 0.8 & 0.78 & 0.85 & 0.81 \\
long illness duration POSITIVE & 0.72 & 0.76 & \textbf{0.74} & 0.79 & 0.64 & 0.7 & 0.7 & 0.69 & 0.69 \\
severe illness POSITIVE & 0.45 & 0.33 & 0.38 & 0.47 & 0.39 & \textbf{0.43} & 0.34 & 0.22 & 0.26 \\
suicidality POSITIVE & 0.61 & 0.64 & 0.63 & 0.69 & 0.61 & \textbf{0.65} & 0.48 & 0.63 & 0.55  \\
multiple hospitalizations POSITIVE & 0.8 & 0.84 & \textbf{0.82} & 0.76 & 0.88 & 0.81 & 0.66 & 0.75 & 0.7 \\
recurrent episodes POSITIVE & 0.63 & 0.62 & 0.63 & 0.65 & 0.73 & \textbf{0.69} & 0.55 & 0.65 & 0.59 \\
improvement POSITIVE & 0.58 & 0.64 & \textbf{0.61} & 0.55 & 0.57 & 0.56 & 0.47 & 0.58 & 0.52 \\
physical comorbidity POSITIVE & 0.72 & 0.76 & \textbf{0.74} & 0.64 & 0.83 & 0.73 & 0.72 & 0.8 & 0.76 \\
mental comorbidity POSITIVE & 0.81 & 0.75 & \textbf{0.78} & 0.78 & 0.72 & 0.74 & 0.81 & 0.72 & 0.76 \\
substance abuse POSITIVE & 0.74 & 0.64 & 0.68 & 0.71 & 0.83 & \textbf{0.77} & 0.67 & 0.78 & 0.72  \\
anhedonia POSITIVE & 0.58 & 0.67 & 0.62 & 0.62 & 0.8 & \textbf{0.7} & 0.59 & 0.7 & 0.64 \\
illness early onset POSITIVE & 0.65 & 0.63 & 0.64 & 0.67 & 0.75 & \textbf{0.7} & 0.61 & 0.55 & 0.58 \\
multiple antidepressants POSITIVE & 0.82 & 0.78 & 0.8 & 0.79 & 0.83 & \textbf{0.81} & 0.7 & 0.79 & 0.74 \\
antidepressant dosage increase POSITIVE & 0.73 & 0.86 & 0.79 & 0.82 & 0.78 & \textbf{0.8} & 0.68 & 0.65 & 0.66 \\
multiple psychotherapies POSITIVE & 0.64 & 0.75 & 0.69 & 0.73 & 0.72 & \textbf{0.72} & 0.66 & 0.8 & 0.72 \\
side effects POSITIVE & 0.81 & 0.71 & \textbf{0.76} & 0.74 & 0.74 & 0.74 & 0.67 & 0.68 & 0.68 \\
non adherence POSITIVE & 0.75 & 0.66 & 0.7 & 0.76 & 0.77 & \textbf{0.76} & 0.68 & 0.68 & 0.68 \\
\midrule
older age NEGATIVE & 0.44 & 0.55 & \textbf{0.49} & 0.58 & 0.41 & 0.48 & 0.49 & 0.39 & 0.43 \\
family member mental disorder NEGATIVE & 0.71 & 0.62 & 0.66 & 0.72 & 0.71 & \textbf{0.72} & 0.62 & 0.5 & 0.55 \\
abuse NEGATIVE & 0.54 & 0.76 & 0.63 & 0.6 & 0.8 & \textbf{0.69} & 0.55 & 0.7 & 0.62 \\
childhood abuse NEGATIVE & 0.73 & 0.74 & \textbf{0.74} & 0.64 & 0.69 & 0.66 & 0.53 & 0.49 & 0.51 \\
long illness duration NEGATIVE & 0.49 & 0.4 & 0.44 & 0.58 & 0.43 & \textbf{0.5} & 0.42 & 0.35 & 0.38 \\
severe illness NEGATIVE & 0.37 & 0.31 & 0.34 & 0.62 & 0.41 & \textbf{0.49} & 0.31 & 0.41 & 0.35 \\
suicidality NEGATIVE & 0.68 & 0.85 & \textbf{0.75} & 0.61 & 0.77 & 0.68 & 0.64 & 0.71 & 0.67 \\
multiple hospitalizations NEGATIVE & 0.76 & 0.83 & \textbf{0.8} & 0.77 & 0.79 & 0.78 & 0.72 & 0.68 & 0.7 \\
recurrent episodes NEGATIVE & 0.75 & 0.48 & 0.59 & 0.76 & 0.49 & \textbf{0.6} & 0.58 & 0.39 & 0.46 \\
improvement NEGATIVE & 0.42 & 0.42 & \textbf{0.42} & 0.35 & 0.49 & 0.4 & 0.34 & 0.26 & 0.29 \\
physical comorbidity NEGATIVE & 0.66 & 0.71 & 0.68 & 0.67 & 0.68 & \textbf{0.68} & 0.59 & 0.61 & 0.6 \\
mental comorbidity NEGATIVE & 0.65 & 0.68 & 0.67 & 0.72 & 0.65 & \textbf{0.68} & 0.5 & 0.56 & 0.53 \\
substance abuse NEGATIVE & 0.7 & 0.91 & \textbf{0.79} & 0.69 & 0.89 & 0.78 & 0.62 & 0.75 & 0.68 \\
anhedonia NEGATIVE & 0.55 & 0.3 & 0.39 & 0.61 & 0.34 & \textbf{0.44} & 0.56 & 0.3 & 0.39 \\
illness early onset NEGATIVE & 0.7 & 0.57 & \textbf{0.63} & 0.75 & 0.53 & 0.62 & 0.53 & 0.5 & 0.51 \\
multiple antidepressants NEGATIVE & 0.32 & 0.2 & 0.25 & 0.59 & 0.37 & \textbf{0.45} & 0.37 & 0.3 & 0.33 \\
antidepressant dosage increase NEGATIVE & 0.56 & 0.57 & 0.56 & 0.54 & 0.59 & \textbf{0.56} & 0.53 & 0.41 & 0.46 \\
multiple psychotherapies NEGATIVE & 0.58 & 0.52 & \textbf{0.55} & 0.63 & 0.37 & 0.47 & 0.47 & 0.2 & 0.28 \\
side effects NEGATIVE & 0.58 & 0.67 & \textbf{0.62} & 0.58 & 0.64 & 0.61 & 0.56 & 0.37 & 0.44 \\
non adherence NEGATIVE & 0.39 & 0.43 & \textbf{0.41} & 0.37 & 0.44 & 0.4 & 0.33 & 0.42 & 0.37 \\
NO ANNOTATION & 0.81 & 0.79 & 0.8 & 0.84 & 0.77 & 0.8 & 0.8 & 0.81 & \textbf{0.81} \\
\midrule
\textbf{POSITIVE} & 0.68 & 0.68 & 0.68 & 0.69 & 0.73 & \textbf{0.70} & 0.64 & 0.67 & 0.64 \\

\textbf{NEGATIVE} & 0.59 & 0.59 & 0.58 & 0.63 & 0.58 & \textbf{0.59} & 0.53 & 0.49 & 0.50 \\ 
\midrule

\textbf{PATIENT} & 0.63 & 0.66 & 0.63 & 0.64 & 0.70 & \textbf{0.65} & 0.63 & 0.59 & 0.59 \\
\textbf{ILLNESS}  & 0.64 & 0.63 & 0.63 & 0.66 & 0.64 & \textbf{0.64} & 0.57 & 0.59 & 0.57 \\ 
\textbf{TREATMENT}  & 0.62 & 0.62 & 0.61 & 0.65 & 0.62 & \textbf{0.63} & 0.58 & 0.52 & 0.54 \\
\midrule
\textbf{All} & 0.63 & 0.63 & 0.63 & 0.66 & 0.65 & \textbf{0.65} & 0.58 & 0.57 & 0.57 \\
\bottomrule
\end{tabular}
\caption{Precision, recall and macro averaged F1 for each model and factor (synthetic data). \textit{best in bold}}
\label{results}
\end{table*}

\begin{table*}[htbp]
\small
\centering
\begin{tabular}{lrrrr}
\toprule
Class & Precision & Recall & F1 Score & N \\
\midrule
NO ANNOTATION &  0.59 & 0.34 & 0.43 & 99 \\
anhedonia POSITIVE & 0.76 & 0.59  & 0.67 & 27 \\
antidepressant dosage increase POSITIVE & 0.31 & 0.83 & 0.45 & 12\\
abuse POSITIVE & 0.94 & 0.80 & 0.86 & 20 \\
family member mental disorder POSITIVE & 0.61 & 0.92 & 0.73 & 12 \\
illness early onset POSITIVE & 0.50 &  1.00 & 0.67 & 8 \\
improvement POSITIVE & 0.68 & 0.42 & 0.52 & 31\\
long illness duration POSITIVE & 1.00 & 0.80 & 0.89 & 5 \\
mental comorbidity POSITIVE & 0.43 & 0.38 & 0.40 & 8 \\
physical comorbidity POSITIVE & 0.50 & 0.67 & 0.57 & 9 \\
multiple antidepressants POSITIVE & 0.75 & 0.62 & 0.68 & 29 \\
multiple psychotherapies POSITIVE & 0.38 & 0.38 & 0.38 & 13 \\
non adherence POSITIVE & 0.50 & 0.19 & 0.27 & 16 \\
recurrent episodes POSITIVE & 1.00 & 0.90 & 0.95 & 10 \\
severe illness POSITIVE & 0.55 & 1.00 & 0.71 & 6 \\
side effects POSITIVE & 0.79 & 0.44 & 0.57 & 34 \\
substance abuse POSITIVE & 0.71 & 0.62 & 0.67 & 13 \\
suicidality POSITIVE & 0.80 & 0.75 & 0.77 & 32 \\
antidepressant dosage increase NEGATIVE & 0.50 & 0.56 & 0.53 & 16 \\
improvement NEGATIVE & 0.60 & 0.16 & 0.25 & 19 \\
multiple antidepressants NEGATIVE & 0.20 & 1.00 & 0.33 & 3 \\
multiple psychotherapies NEGATIVE & 0.51 & 0.93 & 0.66 & 28 \\
multiple hospitalizations NEGATIVE & 0.75 & 1.00 & 0.86 & 3 \\
non adherence NEGATIVE & 0.12 & 0.40 & 0.19 & 5 \\
severe illness NEGATIVE & 0.64 & 1.00 & 0.78 & 7 \\
side effects NEGATIVE & 0.65 & 0.68 & 0.67 & 19\\
substance abuse NEGATIVE & 0.33 & 0.33 & 0.33 & 3 \\
suicidality NEGATIVE & 0.78 & 0.78 & 0.78 & 9 \\
\midrule
\textbf{POSITIVE} & 0.66 & 0.67 & 0.63 & 285 \\

\textbf{NEGATIVE} & 0.51 & 0.68 & 0.54 & 98 \\ 

\midrule

\textbf{All} & 0.60 & 0.66 & 0.60 & 482 \\

\bottomrule
\end{tabular}
\caption{Precision, recall and macro averaged F1 for each factor (clinical data). Classes with n \textless 2 excluded.}
\label{results_real}
\end{table*}

\begin{table*}[htbp]
\small
\centering
\begin{tabular}{lrrr}
\toprule
Class & Precision & Recall & F1 Score \\
\midrule
abuse POSITIVE & 1.00 & 0.80 & 0.89 \\
family member mental disorder POSITIVE & 0.85 & 0.92 & 0.88 \\
severe illness POSITIVE & 1.00 & 0.67 & 0.80 \\
suicidality POSITIVE & 0.96 & 0.75 & 0.84 \\
\midrule

\textbf{All} & 0.95 & 0.78 & 0.85 \\

\bottomrule
\end{tabular}
\caption{Precision, recall and macro averaged F1 for each factor (clinical data -high confidence classes with 0.5 confidence threshold).}
\label{results_highconf}
\end{table*}

\begin{table*}[htbp]
\small
\centering
\begin{tabular}{lrrr}
\toprule
Class & Precision & Recall & F1 Score \\
\midrule
NO ANNOTATION & 0.59 & 0.34 & 0.43 \\
anhedonia & 0.75 & 0.86 & 0.80 \\
antidepressant dosage increase & 0.48 & 0.86 & 0.62 \\
abuse & 0.95 & 0.90 & 0.93 \\
family member mental disorder & 0.61 & 0.92 & 0.73 \\
illness early onset & 0.33 & 1.00 & 0.50 \\
improvement & 0.88 & 0.42 & 0.57 \\
long illness duration & 0.67 & 0.57 & 0.62 \\
mental comorbidity & 0.38 & 0.38 & 0.38 \\
physical comorbidity & 0.50 & 0.80 & 0.62 \\
multiple antidepressants & 0.79 & 0.97 & 0.87 \\
multiple psychotherapies & 0.64  & 1.00 & 0.78\\
multiple hospitalisations & 0.75 & 0.75 & 0.75 \\
non adherence & 0.55 & 0.57 & 0.56 \\
recurrent episodes & 0.92 & 1.00 & 0.96 \\
severe illness & 0.59 & 1.00 & 0.74 \\
side effects & 0.87 & 0.64 & 0.74 \\
substance abuse & 0.88 & 0.79 & 0.83 \\
suicidality & 0.87 & 0.83 & 0.85 \\

\midrule

\textbf{All} & 0.68 & 0.77 & 0.70 \\

\bottomrule
\end{tabular}
\caption{Precision, recall and macro averaged F1 for each factor (clinical data -non-polarised).}
\label{results_nonpol}
\end{table*}

\section{Example extractions}
Here we show some examples of successful span extractions in synthetic test sentences with (start, end, label) for each extracted span:
\begin{itemize}
    \item \textit{Treatment History: patient already tried multiple antidepressant medications from different classes including SSRIs and SNRIs but did not experience significant improvement}. \textbf{spans: (0, 46, multiple antidepressants POSITIVE); (47, 74, improvement NEGATIVE}) 
    \item \textit{XXXXX has been inpatient twice for mental health treatment due to severity of illness with recurrent episodes of major depressive disorder occurring approximately every 3-4 months.} \textbf{spans: (0, 81,  multiple hospitalizations POSITIVE); (82, 175, recurrent episodes POSITIVE}
    \item \textit{FFFFF reports no family history of mental disorders and denies any history of abuse} \textbf{spans: (0, 51, family member mental disorder NEGATIVE); (52, 83, abuse NEGATIVE)}
    \item \textit{patient reports no significant physical comorbidities but mentions mild anxiety symptoms} \textbf{spans: (0, 53, physical comorbidity NEGATIVE); (54, 88, mental comorbidity POSITIVE)}
    \item \textit{has been on citalopram, fluoxetine and sertraline} \textbf{spans: (0, 57, multiple antidepressants POSITIVE)}
    \item \textit{patient recalls being severely bullied at school} \textbf{spans: (0, 48, abuse POSITIVE)}
    \item \textit{she battles with heroin addiction} \textbf{spans: (0, 33, substance abuse POSITIVE)}
    \item \textit{he denies intent to end his own life} \textbf{spans: (0, 36, suicidality NEGATIVE)}
    \item \textit{she did not suffer any neglect as a child} \textbf{spans: (0, 41, abuse NEGATIVE)}
    \item \textit{he talked about how his friend was bullied} \textbf{spans: (0, 42, NO ANNOTATION)}
    \item\textit{her brother was diagnosed with ptsd} \textbf{spans: (0, 35, family member mental disorder POSITIVE)}
\end{itemize}

%

\section{Discussion}

In the synthetic test dataset, we find that our custom span-level model which uses a variant of Non-Maximum Suppression (NMS) outperforms the simpler token-level model which is standardly used for span extraction.
Additionally, we find that the sentence-level model performs slightly under the span-level model overall. 
We hypothesise that this might be because the model learns to more specifically map labels with the relevant tokens, rather than relying on fuzzier learning over full sentences.

Performance on positive classes is significantly higher than for negative classes. 
This is due to a number of reasons. First, despite our additional prompting for negative factors exclusively, there is still an imbalance with more positive than negative labels. Second, it is a well-known fact that language models struggle with negation \cite{ettinger2020bert}. Finally, negation will often be present in the sentence but not necessarily in each negated span, even if the scope of the negation encompasses the span, e.g., \textit{The patient denies suicidality [suicidality NEGATIVE] and substance abuse [substance abuse NEGATIVE]}, where the text of the second span does not contain an explicit negation. While each span's tokens contextual embeddings should have some signal which indicates there is a negation somewhere in the sentence, it might not be strong enough compared to classifying a span which contains an explicit negation token. 

While average F1 on real data is 0.60, our model trained exclusively on synthetic data already has practical clinical use, as it can be used out of the box with a confidence threshold of 0.5 to extract abuse, family disorder, severe illness and suicidality with 0.85 F1 and 0.95 precision, a performance comparable to \citet{kulkarni2024enhancing}, who used real data annotated manually in a costly and time-consuming way to train a model to extract two clinical factors (suicidality and anhedonia). This is despite the high syntactic and semantic variability among spans expressing these factors. For example, our model is able to identify spans mentioning events as varied as emotional neglect, violence or bullying as abuse, and a wide range of combination of family member and various conditions (brother, aunt, schizophrenia, bipolar disorder, etc.) as a family history of mental disorder. Obtaining such high performance with a model trained on synthetic data only shows this is a promising direction of research for cost-efficient AI applications in healthcare. Indeed, the cost of producing the synthetic training data used in this study was under £10, versus the thousands of pounds required for compensating expert annotators.

We believe there are several reasons the model fails to achieve higher average performance across all factors. First, many classes are very close to one another, for example long duration and early onset, classes which mention antidepressants, physical comorbidities and side effects, etc. It is no coincidence that the model performs best on classes which are most distinctive (abuse, family disorder, suicidality).
Secondly, many classes are highly subjective, e.g., early onset (how early?), long duration (how long?), substance abuse (how much consumption?) and spans are often ambivalent (e.g., `some improvement then worsening'; `some side effects then none', etc.)
Finally, many negative classes are not well defined or consistent, e.g, negative multiple psychotherapies (mentions only one therapy? any span mentioning therapy?), negative multiple antidepressants (any span mentioning a single antidepressant?), negative multiple hospitalisations? (having been hospitalised once? Never?), negative anhedonia (any span mentioning subject performing activities?), negative severe illness (moderate illness? mild symptoms?), etc. In view of these challenges, it is likely that training a model to extract a wide range of factors requires some real annotated clinical data, however impressively high performance can already be achieved on a subset of factors by exclusively training on synthetic data.

\section{Future work}
The paradigm we used could be extended and scaled to other phenotypes with a different set of risk/prognostic factors or clinical features in the future. Future research could also investigate whether using a model which has been domain pretrained on mental health data (such as MentalBERT, \citep{DBLP:journals/corr/abs-2110-15621} would improve performance. Works such as \cite{guevara2024} suggest that a sequence-to-sequence model such as T5 could achieve even better performance than a BERT-based classifier model. Using GPT4 instead of GPT3.5 could help generate synthetic data with reduced noise and more accurate labelling of negative factors. Finally, an optimal weighting scheme for the extracted factors which would best allow identification of difficult-to-treat depression could be developed in consultation with clinicians. 

\section{Conclusion}
The goal of this study was to train a model to extract spans which contain factors associated with the syndrome of difficult-to-treat depression. To achieve this, we generated annotated synthetic clinical notes with both positive and negative factors of interest using a Large Language Model (GPT3.5-turbo) and subsequently trained various BERT-based classifier models (sentence, token and span level) to extract factors. We show it is possible to obtain good performance on real clinical data on a set of as many of 20 different factors, and high performance on a subset of clinically-relevant factors by training exclusively on LLM-generated synthetic data. 


\section {Acknowledgements}

I.L., A.K. and D.W.J. were supported in part by the NIHR AI Award for Health and Social Care (AI-AWARD02183), A.K. by a research grant from GlaxoSmithKline. The views expressed are those of the authors and not necessarily those of the UK National Health Service, the NIHR or the UK Department of Health. This study was supported by CRIS Powered by Akrivia Health, using data, systems and support from the NIHR Oxford Health Biomedical Research Centre (BRC-1215-20005) Research Informatics Team. We would also like to acknowledge the work and support of the Oxford Research Informatics Team: Tanya Smith, Research Informatics Manager, Adam Pill, Suzanne Fisher, Research Informatics Systems Analysts and Lulu Kane Research Informatics Administrator.


\newpage

\bibliography{custom}
\bibliographystyle{acl_natbib}

\onecolumn
\newpage
\appendix
\section{Appendix A}
\label{AppendixA}
\begin{table}[h!]
\centering
\begin{tabular}{|m{17cm}|} 
 \hline

Patient presentation: 
The patient is a 45-year old female [PATIENT\_FACTOR(POSITIVE):older\_age] who presents with a history of mental illness in her family, with her sister suffering from bipolar disorder [PATIENT\_FACTOR(POSITIVE):family\_member\_mental\_disorder]. She reports a childhood marked by abuse, specifically physical abuse from her father [PATIENT\_FACTOR(POSITIVE):childhood\_abuse]. The patient further elaborates that she experienced emotional neglect from her mother for several years [PATIENT\_FACTOR(POSITIVE):childhood\_abuse].

Illness history: 
The patient's psychiatric illness began at the age of 18 with the onset of major depressive episodes. Over the past 27 years [ILLNESS\_FACTOR(POSITIVE):long\_illness\_duration], she has experienced multiple episodes of severe depression, including thoughts of ending her life [ILLNESS\_FACTOR(POSITIVE):suicidality]. Despite the severity of her illness, she has never been hospitalized [ILLNESS\_FACTOR(NEGATIVE):multiple\_hospitalizations]. The patient has a history of recurrent depressive episodes [ILLNESS\_FACTOR(POSITIVE):recurrent\_episodes].

Treatment history: 
The patient's treatment history has involved a variety of interventions. She has been prescribed multiple antidepressant medications over the years [TREATMENT\_FACTOR(POSITIVE):multiple\_antidepressants] at varying dosages, including increases in dose [TREATMENT\_FACTOR(POSITIVE):antidepressant\_dosage\_increase]. However, she has experienced side effects such as weight gain, sedation, and sexual dysfunction, prompting changes in medication regimens [TREATMENT\_FACTOR(POSITIVE):side\_effects]. The patient has also been engaged in multiple psychotherapies [TREATMENT\_FACTOR(POSITIVE):multiple\_psychotherapies].

Current presentation: 
During today's session, the patient reports some improvement in her symptoms. She notes a decrease in depressive symptoms such as sadness and hopelessness. However, she still experiences anhedonia [ILLNESS\_FACTOR(POSITIVE):anhedonia] and struggles with maintaining positive relationships. There is evidence of physical comorbidity as the patient shares that she was recently diagnosed with diabetes [ILLNESS\_FACTOR(POSITIVE):physical\_comorbidity]. Additionally, she has a comorbid diagnosis of generalized anxiety disorder [ILLNESS\_FACTOR(POSITIVE):mental\_comorbidity]. The patient denies any current substance abuse [ILLNESS\_FACTOR(NEGATIVE):substance\_abuse].

Discussion and plan: 
The patient's depressive symptoms seem to have an early onset [ILLNESS\_FACTOR(POSITIVE):illness\_early\_onset], starting at 18 years old. Her history of multiple episodes of depression, suicidal ideation, and non-adherence to medication regimens suggest a severe and chronic illness course [ILLNESS\_FACTOR(POSITIVE):severe\_illness, ILLNESS\_FACTOR(POSITIVE):recurrent\_episodes, ILLNESS\_FACTOR(POSITIVE):non\_adherence]. We will continue to monitor her progress and consider further adjustments to her medication regimen based on her response and any side effects. Therapy sessions will focus on enhancing coping skills, reducing anhedonia, and improving interpersonal relationships. We will also explore strategies to address the impact of childhood abuse on her current mental health. Emergency contact information will be reviewed, emphasizing the importance of seeking help during times of intense distress or suicidal thoughts. Follow-up appointments will be scheduled to assess treatment response and assess any additional needs.\\
\hline
\end{tabular}
\caption{Example synthetic annotated note}
\end{table}

\section{Appendix B}
\label{AppendixB}

\begin{table}

\begin{tabular}{|m{17.5cm}|} 
 \hline
  """
Create a narrative psychiatric clinical note (about 500 words) and annotate sentences with\\

PATIENT-RELATED factors (older age, family member mental disorder, abuse, childhood abuse), 
ILLNESS-RELATED factors (long illness duration, severe illness, suicidality, multiple hospitalizations, recurrent episodes, improvement, physical comorbidity, mental comorbidity, substance abuse, anhedonia, illness early onset) 
and TREATMENT-RELATED factors (multiple antidepressants, antidepressant dosage increase, multiple psychotherapies, side effects, non adherence).
\\
Make sure to annotate all relevant text for each factor.
The annotation must indicate if the evidence is POSITIVE (presence of factor) or NEGATIVE (absence of factor) in the exact same format as the following examples:
\\
"The patient is a 32-year old male [PATIENT\_FACTOR(NEGATIVE):older\_age]."
"The patient reports a history of mental illness in her family, with her brother suffering from generalised anxiety 
[PATIENT\_FACTOR(POSITIVE):family\_member\_mental\_disorder]
and that she visits him regularly.";
\\
"She experienced childhood abuse, suffering emotional neglect from her mother for several years [PATIENT\_FACTOR(POSITIVE):childhood\_abuse].";
\\
Annotate each factor separately within each sentence after the relevant text, e.g.: \\
"The dosage was gradually increased [TREATMENT\_FACTOR(POSITIVE):antidepressant\_dosage\_increase] until severe gastrointestinal distress emerged [TREATMENT\_FACTOR(POSITIVE):side\_effects] leading to a review of current approach." \\
"She experienced multiple periods of depression [ILLNESS\_FACTOR(POSITIVE):recurrent\_episodes] but was never hospitalized [ILLNESS\_FACTOR(NEGATIVE):multiple\_hospitalizations] as she opposed it repeatedly."; \\
"He was treated with various antidepressants [ILLNESS\_FACTOR(POSITIVE):multiple\_antidepressants] and experienced relapses upon discontinuation [ILLNESS\_FACTOR(POSITIVE):non\_adherence] which was due to a general distrust in medications." \\
Do not annotate the plan.
""" \\
 \hline
\end{tabular}
\caption{ChatGPT prompt}
\label{table:3}
\end{table}

\newpage
\section{Appendix C}
\label{AppendixC}

\begin{table}
\begin{tabular}{|m{17.5cm}|} 
\hline
"""
Create a narrative psychiatric clinical note (about 500 words) and annotate sentences with negative examples of the following factors:\\
PATIENT-RELATED factors (older age, family member mental disorder, abuse, childhood abuse), 
ILLNESS-RELATED factors (long illness duration, severe illness, suicidality, multiple hospitalizations, recurrent episodes, improvement, physical comorbidity, mental comorbidity, substance abuse, anhedonia, illness early onset) 
and TREATMENT-RELATED factors (multiple antidepressants, antidepressant dosage increase, multiple psychotherapies, side effects, non adherence). \\
Make sure to annotate all relevant text for each factor.
The annotation must indicate that the evidence is NEGATIVE (absence of factor) in the exact same format as the following examples: \\
"The patient is a 32-year old male [PATIENT\_FACTOR(NEGATIVE):older\_age]." \\
"She was never hospitalized [ILLNESS\_FACTOR(NEGATIVE):multiple\_hospitalizations] as she opposed it repeatedly."; \\
Do not annotate the plan.
""" \\
\hline
\end{tabular}
\caption{ChatGPT negative prompt}
\end{table}

\newpage
\section{Appendix D}
\label{AppendixD}
\begin{figure}[h]
\centering
\includegraphics[]{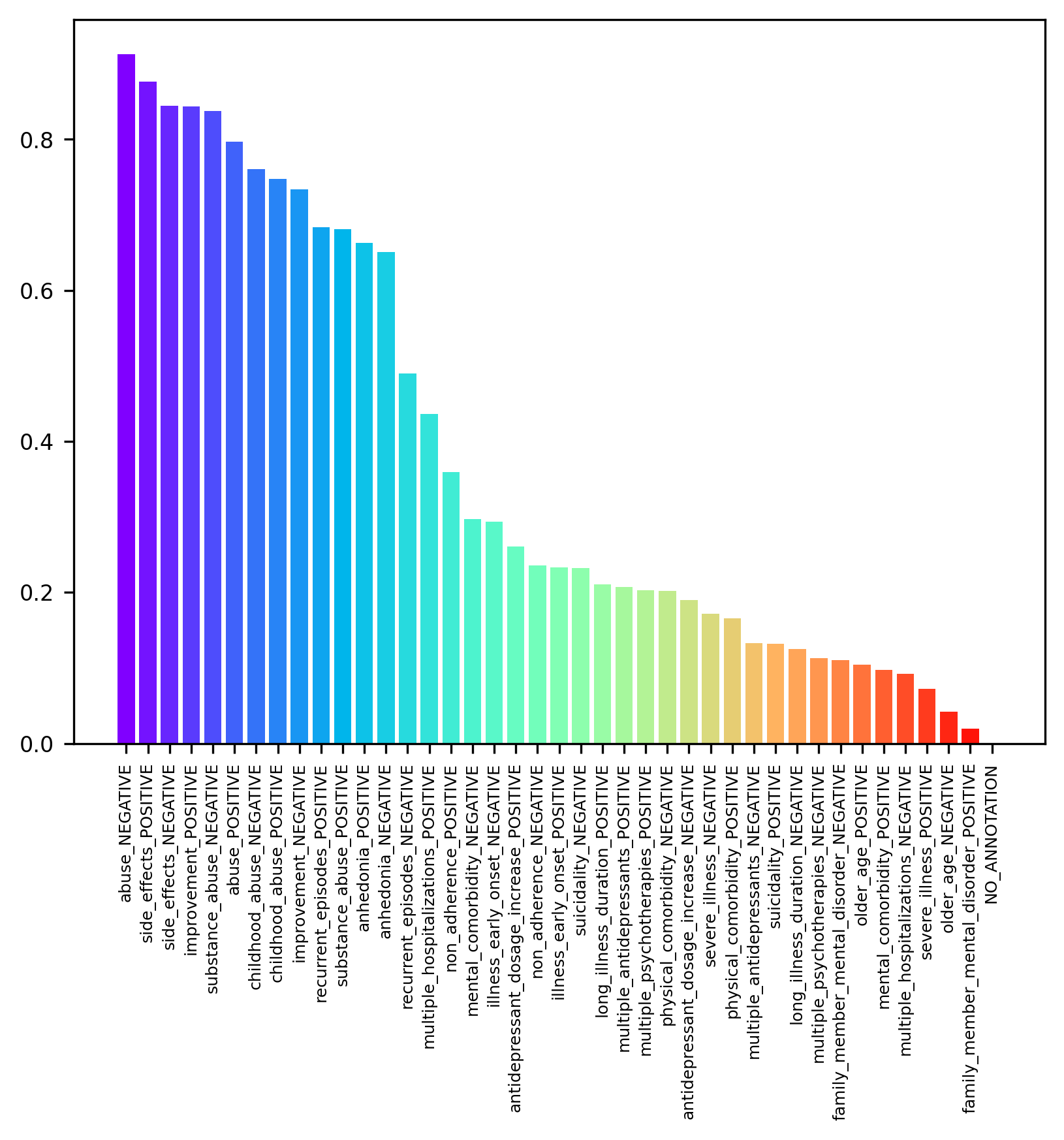}
\caption{Average explicit mentions of all label words.}
\label{explicit}
\end{figure}

\begin{figure}[h]
\centering
\includegraphics[]{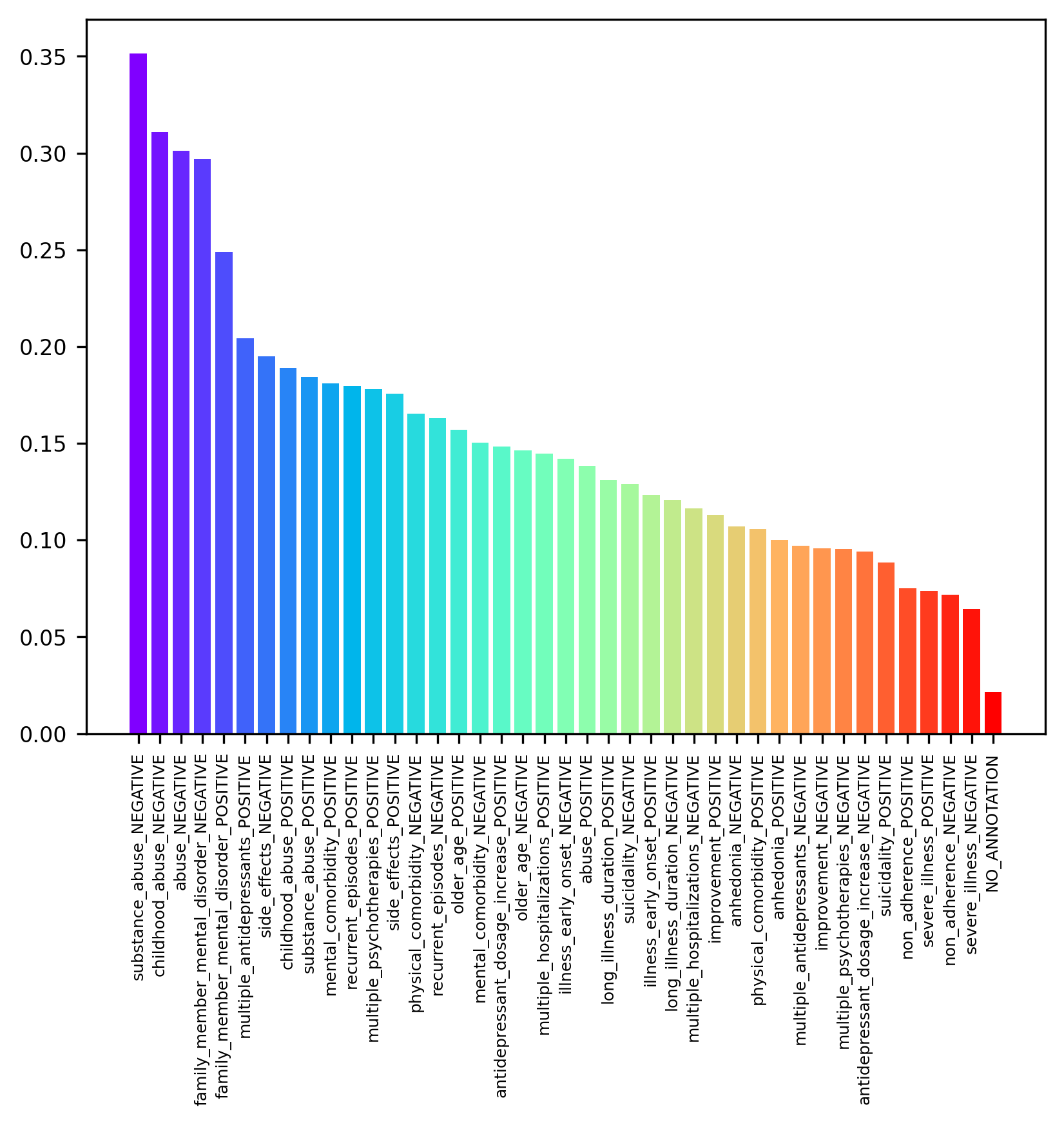}
\caption{Average Jaccard similarities of span pairs.}
\label{jaccard}
\end{figure}

\end{document}